\documentclass[article, shortnames, nojss]{jss}

\usepackage{framed}
\usepackage[utf8]{inputenc} 
\usepackage[T1]{fontenc}    
\usepackage{hyperref}       
\usepackage{url}            
\usepackage{booktabs}       
\usepackage{amsfonts}       
\usepackage{nicefrac}       
\usepackage{microtype}      
\usepackage{xcolor}         
\usepackage{amsmath}
\usepackage{graphicx}
\usepackage{caption}
\usepackage{subcaption}
\usepackage{comment}
\usepackage{listings}
\usepackage{float}
\usepackage{hhline}
\usepackage{soul}
\usepackage{multirow}
\usepackage{tabstackengine}
\usepackage{amsthm}

\usepackage{booktabs}

\newcommand{\softa}{\pkg BoXHED1.0 }

\usepackage{pifont}
\newcommand{\softb}{\pkg BoXHED1.0}
\newcommand{\softc}{\pkg BoXHED2.0 }
\newcommand{\softd}{\pkg BoXHED2.0}
\newcommand{\softe}{\pkg BoXHED }
\newcommand{\softf}{\pkg BoXHED}
\newcommand{\xgb}{\pkg XGBoost }
\newcommand{\xgbb}{\pkg XGBoost}

\newcommand{\blackboost}{\pkg mboost }

\newcommand{\numpy}{\pkg NumPy }

\newcommand{\pandas}{\pkg pandas }

\newcommand{\cv}{{\ttfamily cv()} }
\newcommand{\cvb}{{\ttfamily cv()}}

\newcommand{\tstart}{{\ttfamily t\_start} }
\newcommand{\tstartb}{{\ttfamily t\_start}}

\newcommand{\tendb}{{\ttfamily t\_end}}

\newcommand{\deltab}{{\ttfamily delta}}

\newcommand{\idb}{{\ttfamily ID}}

\newcommand{\maxdepthb}{{\ttfamily max\_depth}}

\newcommand{\nestimatorsb}{{\ttfamily n\_estimators}}

\usepackage{etoolbox}

\AtBeginEnvironment{array}

{\setlength{\arraycolsep}{3.8pt}}

\usepackage{longtable}

\author{Arash Pakbin\\
	Texas A\&M University\\
  \And
	Xiaochen Wang\\
	Yale University\\
	\And
	Bobak J. Mortazavi\\
	Texas A\&M University\\
	\And
	Donald K.K. Lee\\
	Emory University\\
        }
\Plainauthor{Arash Pakbin, Xiaochen Wang, Bobak J. Mortazavi, Donald K.K. Lee}

\title{\softd: Scalable Boosting of Dynamic Survival Analysis}
\Plaintitle{BoXHED2.0: Scalable Boosting of Dynamic Survival Analysis}
\Shorttitle{\softd: Boosted Hazard Learning}

\Abstract{
Modern applications of survival analysis increasingly involve time-dependent covariates. 
The \proglang{Python} package \softc is a tree-boosted hazard estimator that is fully nonparametric, and is applicable to survival settings far more general than right-censoring, including recurring events and competing risks.
\softc is also scalable to the point of being on the same order of speed as parametric boosted survival models, in part because its core is written in \proglang{C++} and it also supports the use of GPUs and multicore CPUs. \softc is available from PyPI and also from \href{http://github.com/BoXHED}{www.github.com/BoXHED}.
}

\Plainkeywords{gradient boosting, nonparametric estimation, survival analysis, hazard estimation, survivor function, competing risks, time-varying covariates, Python}

\Address{Arash Pakbin\\
	Department of Computer Science \& Engineering\\
	Texas A\&M University \\
        College Station TX, USA\\
        E-mail: \email{a.pakbin@tamu.edu}\\\\
	Xiaochen Wang\\
	Department of Biostatistics\\
	Yale University\\
        New Haven CT, USA\\
        E-mail: \email{xcwang11@gmail.com}\\\\
	Bobak J. Mortazavi\\
	Department of Computer Science \& Engineering\\
	Texas A\&M University\\
        College Station TX, USA\\
        E-mail: \email{bobakm@tamu.edu}\\\\
	Donald K.K. Lee\\
	Goizueta Business School and Dept of Biostatistics \& Bioinformatics\\
	Emory University\\
        Atlanta GA, USA\\
        E-mail: \email{donald.lee@emory.edu}
        }

\begin{document}



\section{Introduction}\label{sec:intro}

Survival analysis is concerned with analyzing the time $T$ to an event of interest,
and the fundamental quantity of interest is the hazard function $\lambda(t,X(t))$. This is informally the probability of $T\in[t,t+dt)$ given that the event has not yet occurred by $t$. Here, $X(t)\in\mathbb{R}^p$ denotes the predictable covariate process which can vary over time. We may think of the hazard as the survival analogue to the probability density function. For the special case where $X(t)=X$ is time-static, there also exists an analogue to the cumulative distribution called the survivor function $S(t|X) = \mathbb{P}(T>t|X)$, which can be derived from the hazard via
\[
S(t|X) = \exp\left(-\int_0^t \lambda(u,X)du\right).
\]
In general, if $X(t)$ is time-varying, the survivor function $S(t|X)$ is undefined because we do not know the future path of $\{X(u)\}_{0<u\le t}$. Also, if the event of interest can recur (e.g., cancer relapse), the survivor function tells us nothing about events subsequent to the first. On the other hand, the hazard is well defined in both situations and represents the realtime risk of the event (re)occurring. For example, a patient's risk of stroke at a given point in time depends on factors such as heart rate and previous stroke history.
The hazard as a function of these time-varying factors quantifies the patient's realtime risk of stroke.

Furthermore, in a competing risks setting where the occurrence of one type of event precludes the occurrence of the others (e.g., employee termination vs. resignation), the fundamental quantities that are identifiable are the cause-specific hazards \citep{andersen2002competing}. The cause-specific hazard for each event type can be estimated by treating the occurrence of the other types as censoring, and then applying hazard estimation procedures such as the one introduced here. This again illustrates the unifying role played by the hazard function across a variety of survival settings.

This paper presents the \proglang{Python} library \softd, a nonparametric hazard estimator that can handle high-dimensional, time-dependent covariates. It is a novel tree-based implementation of the gradient-boosted estimator proposed in \citep{lee2021boosted} for generic learners. The previous version \softa \citep{wang2020boxhed} can only deal with right-censored data. \softc is redesigned using a counting process framework to accommodate more general censoring schemes like those described above. It is also far more scalable than \softa because its core engine is implemented in \proglang{C++} and also because of a novel data preprocessing step. \softc also provides built-in functionality for obtaining the survivor curve $S(t|X)$ from the estimated hazard function when the covariates are time-static.

The rest of the paper is organized as follows. Section~\ref{sec:lit} reviews related survival boosting packages. Section~\ref{sec:alg} describes the \softc implementation. Section~\ref{sec:software} introduces the \softc software and its \proglang{Python} interface. The scalability of \softc is studied in Section~\ref{sec:computational_complexity}. Section~\ref{sec:boxhed_in_practice} provides an example of a real life situation where the generality of the censoring schemes supported by \softc is needed.


In \proglang{Python}, \softc is available via {\ttfamily pip} and also via pre-compiled packages. Installation instructions and a tutorial for how to run \softc can be found at \href{http://github.com/BoXHED}{github.com/BoXHED}.

\section{Related packages}\label{sec:lit}

There are a number of survival boosting packages for the case of time-static covariates. The \proglang{R} package {\pkg mboost} can be used to fit boosted parametric accelerated failure time (AFT) models \citep{schmidAFT,hothorn2010model}. Boosted semiparametric Cox models can also be fit with {\pkg mboost} and also with the {\pkg gbm} package \citep{ridgeway} in \proglang{R}. In \proglang{Python} this can be achieved with the \xgb package \citep{xgboost}. For further flexibility, the \proglang{R} package {\pkg tbm} provides a boosting procedure for flexible transformation models of parametric families \citep{hothorn2017transformationforests,hothorn2020transformation}.

Machine learning work on the general time-dependent survival setting is much more recent. To our knowledge \softc is the only nonparametric boosting package that extends beyond right-censored survival data. The closest package is \softb, which is based on a special case of the \softc implementation that deals only with right-censored data. 
\softa is entirely implemented in \proglang{Python}, whereas the core engine of \softc is written in \proglang{C++} and supports multicore CPU and GPU training. \softc employs a novel data preprocessing step that removes the need for explicit integral evaluations, which is a major bottleneck in \softb. These innovations make \softc orders of magnitude more scalable than \softb.

\color{black}

\section{Implementation details}\label{sec:alg}

\subsection{Overview of survival setting}\label{subsec:setting2.0}
\softc adopts the Aalen intensity model \citep{aalen1978} which encompasses a vast range of survival settings beyond right-censoring. Under the setting considered, the probability of an event occurring in $[t,t+dt)$ is 
\begin{equation}\label{eq:intensity}
\lambda(t,X(t))Y(t)dt,
\end{equation}
where $Y(t)\in \{0,1\}$ is a predictable process indicating whether the subject is at-risk of experiencing an event during $[t,t+dt)$. Further, let $N(t)$ be the cumulative number of events that has occurred by $t$, and $N(t-)$ the count that excludes any potential event occurring exactly at time $t$. 
Some special cases of the setting \eqref{eq:intensity} include:
\begin{itemize}
    \item Right-censored and non-recurring events: $Y(t) = I(t\le C, N(t-)<1)$. Here, $C$ is the right-censoring time.
    \begin{itemize}
        \item Cause-specific hazards in the competing risks setting can be estimated as a further special case of this \citep{andersen2002competing}.
    \end{itemize}
    \item Recurring events (up to $N_{\max}$ times): $Y(t) = I(N(t-)<N_{\max} )$.
    \item Left-truncation and right-censoring: The likelihood for this coincides with the one obtained from setting $Y(t)=I(L\le t\le C, N(t-)<1)$, where $L$ is the left-truncation time. Hence, the same computational procedure applies.
    
    \item Cancer relapse: $Y(t)=1$ whenever a patient is in remission and hence is at risk of relapse, and $Y(t)=0$ if the patient is in relapse.
\end{itemize}
The event history of a subject up to time $\tau$ is captured by the functional data point $\{X(t),Y(t), N(t)\}_{t\le\tau}$. For recurring events, $X(t)$ might include variables like time since the last event and/or the number of past events $N(t-)$.\footnote{The choice of $N(t-)$ rather than $N(t)$ ensures that $X(t)$ remains a predictable process.}

Under \eqref{eq:intensity}, the process generating the first event time can be described as follows: Conditional on the first event having not occurred by $t$, the probability that it happens in $[t,t+dt)$ equals $\lambda(t,X(t))Y(t)dt$. Thus the likelihood of observing the first event at $T_{(1)}$ follows a sequence of coin flips at $t = 0,dt,2dt,\cdots$, i.e.,
\begin{gather*}
\{1-\lambda(0,X(0))Y(0)dt\} \times \{1-\lambda(dt,X(dt))Y(dt)dt\} \times \cdots \times \lambda(T_{(1)},X(T_{(1)})) \\
\xrightarrow[dt\downarrow 0]{}
\exp\left\{-\int_0^{T_{(1)}} Y(t)\lambda(t,X(t))dt \right\}\lambda(T_{(1)},X(T_{(1)})) \\
= \exp\left\{-\int_0^{T_{(1)}} Y(t)\lambda(t,X(t))dt + \log\lambda(T_{(1)},X(T_{(1)})) \right\},
\end{gather*}
where the limit can be understood as a product integral.
Continuing this line of argument for potential subsequent events leads to the following likelihood for all observed events $T_{(1)}, T_{(2)},\cdots$ up to time $\tau$:
\begin{gather*}
\exp\left\{-\int_0^\tau Y(t)\lambda(t,X(t))dt + \sum_k \log\lambda(T_{(k)},X(T_{(k)})) \right\} \\
= \exp\left\{-\int Y(t)\lambda(t,X(t))dt + \int \{\log\lambda(t,X(t))\} dN(t) \right\},
\end{gather*}
where we drop mention of the integral ranges since we can set $Y(t)=0$ and $dN(t)=0$ for $t>\tau$. Note that the likelihood also captures the case where censoring is present. For example, if the subject was censored at time $C$ before any events have occurred, we have $Y(t)=I(t\le C)$ and $N(t)_{t\le C}=0$. The likelihood then reduces to the more familiar $\exp\left\{-\int_0^C \lambda(t,X(t))dt\right\}$.

Thus letting $F(t,x) := \log \lambda(t,x)$ be the log-hazard function and given $n$ functional data samples
\begin{equation}\label{eq:2.0data}
\{X_i(t),Y_i(t), N_i(t)\}_t,
\end{equation}
we define the \textit{likelihood risk} as the negative log-likelihood
\begin{equation}\label{eq:loglik}
R_n(F) = \sum_{i=1}^{n} \left\{ \int Y_i(t)e^{F(t,X_{i}(t))}dt - \int F(t,X_i(t)) dN_i(t)\right\}.
\end{equation}
By the likelihood principle, a function $\hat F$ that minimizes $R(F)$ provides a candidate for the log-hazard estimator, with $\hat \lambda = e^{\hat F}$ being the corresponding hazard estimator. The consistency for such a hazard estimator obtained from gradient boosting is formally established in \citet{lee2021boosted} for arbitrary learners.\footnote{Under mild identifiability conditions the estimator converges in probability to the tree ensemble that best approximates the true hazard.} The \softc estimator is a novel implementation that employs tree learners in particular.


\subsection{BoXHED2.0}
We aim to construct a tree ensemble for the log-hazard estimator
\[
F_M(t,x) = F_0 + \nu\sum_{m=0}^{M-1} g_{m}(t,x)
\]
that iteratively reduces \eqref{eq:loglik}, i.e., the boosted nonparametric maximum likelihood estimator (MLE). Here, $g_0(t,x),\cdots,g_{M-1}(t,x)$ are tree learners of limited depth. The initial guess $F_0 = \log\frac{\sum_i N_i(\infty)}{\sum_i \int Y_i(t)dt}$ is the best minimizing constant for \eqref{eq:loglik}, $M$ is the number of boosting iterations, and $\nu\ll 1$ is the learning rate. The \softc estimator is given by
\begin{equation}\label{eq:boostedhaz}
\hat\lambda(t,x) = e^{F_M(t,x)}.
\end{equation}
In traditional gradient boosting \citep{friedman}, at the $m$-th iteration the tree $g_m$ is constructed to approximate the negative gradient function of the risk at $F=F_m$, the direction of steepest descent.
More recent implementations of boosting such as \xgb constructs $g_m$ in a more targeted manner, growing it to directly minimize the second order Taylor approximation to the risk. \softc takes this one step further by growing trees to directly minimize the exact form of \eqref{eq:loglik}, resulting in even more targeted risk reduction:\footnote{If $R_n(F)$ can be reduced by moving in the direction of a tree learner $g$, then $g$ is necessarily aligned to the negative gradient function of $R_n(F)$ because the risk is convex \citep{lee2021boosted}. As shown in \citet{lee2021boosted}, alignment is needed for consistency, and directly fitting the learner to the negative gradient is just one of many possible ways to achieve this.} Starting with a tree learner $g_{m,0}$ of depth zero (the root node being the only leaf node), we split each leaf node to maximally reduce $R_n(F)$ and repeat the process. Thus the intermediate tree of depth $l$ is\footnote{Similar to \xgbb, to obtain a tree of depth $l$, our implementation splits every node above it to obtain $2^l$ terminal nodes.}
\[
g_{m,l}(t,x)=\sum_{\ell=1}^{2^l} c_{m,\ell}I_{B_{m,\ell}}(t,x),
\]
where $B_{m,\ell}$ represents the time-covariate region for the $\ell$-th leaf node of the form
\begin{equation}\label{eq:regions}
B_\ell=\left\{(t,x) : \begin{pmatrix}
    t^{(\ell_0)}<t\leq t^{(\ell_0+1)}\\
    x^{(1,\ell_1)}<x^{(1)}\leq x^{(1,\ell_1+1)}\\
    \vdots\\
    x^{(p,\ell_p)}<x^{(p)}\leq x^{(p,\ell_p+1)}
    \end{pmatrix}\right\},
\end{equation}
and $c_{m,\ell}$ is the value of the tree function in that region. Here, $x^{(k)}$ denotes the $k$-th covariate. To obtain $g_{m,l+1}(t,x)$ from $g_{m,l}(t,x)$, we split each leaf region $B_{m,\ell}$ in $g_{m,l}$ into left and right daughter regions $A_L$ and $A_R$ by either splitting on time $t$ or on one of the covariates $x^{(1)},\cdots,x^{(p)}$. Since the leaf node regions are disjoint, the restriction of $g_{m,l+1}$ to $(t,x)\in B_{m,\ell}$ is
\[
\begin{aligned}
	\left. g_{m,l+1}(t,x)\right|_{B_{m,\ell}} = \gamma_L I_{A_L}(t,x)+\gamma_R I_{A_R}(t,x).
\end{aligned}
\]
The variable or time axis to split on, the location of the split, and also the values of $(\gamma_L,\gamma_R)$ are all chosen to directly minimize $R_n(F_m + g_{m,l+1})$. Since the values $\gamma_L,\gamma_R$ only apply to the subregions $A_L, A_R$ that are inside $B_{m,\ell}$, $R(F_{m}+g_{m,l+1})$ equals
\[
\begin{aligned}
& \sum_{i=1}^n \sum_{k=L,R}\Biggl\{ \int_{(t,X_i(t))\in A_k} Y_i(t)e^{F_{m}(t,X_{i}(t)) + \gamma_k }dt 
 - \int_{(t,X_i(t))\in A_k} \left[F_m(t, X_i(t)) + \gamma_k\right] dN_i(t) \Biggr\}+C \\
= & \sum_{i=1}^n \sum_{k=L,R} \Biggl\{ e^{\gamma_k} \int_{(t,X_i(t))\in A_k} Y_i(t) e^{F_{m}(t,X_{i}(t))} dt - \gamma_k \int_{(t,X_i(t))\in A_k} dN_i(t) \Biggr\} + C' \\
= & \sum_{k=L,R} (e^{\gamma_{k}}U_k-\gamma_{k}V_k)+C'.
\end{aligned}
\]

where $C,C'$ do not depend on $\gamma_L$ or $\gamma_R$, and
\begin{equation}\label{eq:UV}
U_k = \sum_{i=1}^{n}\int_{(t,X_i(t))\in A_k} Y_i(t) e^{F_{m}(t,X_{i}(t))}dt,
\qquad
V_k = \sum_{i=1}^n \int_{(t,X_i(t))\in A_k} dN_i(t).
\end{equation}
Hence $R_n(F_m + g_{m,l+1})$ is minimized by $\gamma_k=\log(V_k/U_k)$. Adopting the convention that $0\times\infty=0$, the drop in risk $R_n(F_m+g_{m,l}) - R_n(F_m+g_{m,l+1})$ from splitting $B_{m,\ell}$ into ($A_L,A_R$) is
\begin{equation}\label{eq:splitscore}
\Pi = (V_L+V_R)\log\left( \frac{U_L+U_R}{V_L+V_R}\right) -V_L\log\left(\frac{U_L}{V_L}\right) - V_R\log\left(\frac{U_R}{V_R}\right).
\end{equation}
Therefore the best split ($A_L,A_R$) of $B_{m,\ell}$ is that which maximizes $\Pi$. Since $\Pi$ does not depend on the other disjoint leaf regions, all $2^l$ leaf nodes of $g_{m,l}$ can be split in parallel.

The quantities \eqref{eq:UV} and \eqref{eq:splitscore} are calculated for every possible split in order to determine the best split for each $B_{m,\ell}$. Note that $V_k$ is the total number of events observed in the region $A_k$ for the $n$ subjects, and may include more than one event per subject due to the possibility of recurring events.

\subsection{Speedup via data preprocessing}\label{subsec:preprocess}
Three factors contribute to \softd's massive scalability. The first is the use of \proglang{C++} for the core calculations, based in part on the \xgb codebase \citep{xgboost}, a highly efficient boosting package for nonfunctional data. The second is explicit parallelization via multicore CPUs and GPUs. The third is a novel data preprocessing step: Recall that the integral $U_k$ in \eqref{eq:UV} must be calculated for every possible split in order to identify the one that maximizes the score \eqref{eq:splitscore}.
The preprocessing step in \softc transforms the functional survival data \eqref{eq:2.0data} in such a way that the required numerical integration comes for free as part of training. Details can be found in the Appendix. It is worth noting that a training set only needs to be preprocessed once, rather than for each time the \softc estimator is fit with a particular set of hyperparameters during the tuning step.

\subsection{Split search using multicore CPUs and GPUs}\label{subsec:quantiles}

For a continuous covariate, traditional boosting implementations typically place candidate split points at every observed covariate value. This takes $\mathcal{O}(n)$ trials to search through all possible splits for one covariate. On the other hand, picking a pre-specified set of quantiles (e.g., every percentile) of the observed values as candidate splits reduces the search time to $\mathcal{O}(1)$.
The default option in \softc is to use 256 quantiles as candidate split points for time and for each covariate.\footnote{For GPU training of \softc models, the resource bottleneck is typically GPU memory. Using 256 candidate splits allows the covariate values to be stored as a byte rather than as a double \citep{lou2013}.} Users may choose to use a different number, or even supply their own candidate split points, but in either case the number of candidate splits may not exceed 256. 
\softc offers two flavours of quantiles: Raw and time-weighted.

\textit{Raw quantiles.} The set of unique values for time and for each covariate are collected, and the quantiles are obtained from this.

\textit{Weighted quantiles.}
In \xgb the risk function is approximated by its second order Taylor expansion, and the Hessian is used as the weight for quantile sketch. For the time-dependent survival setting we propose a much more natural weight, i.e., time. To illustrate, imagine a sample with one subject ($n=1$) whose covariate value is $x=1.3$ for $t\in (0,2]$ and $x=2$ for $t\in (2,3]$. Under the raw quantile setting, $x=1.3$ and $x=2$ are each given a weight of 1/2. However, since twice as much time was spent at $x=1.3$ than at $x=2$, in a weighted setting we give $x=1.3$ a weight of $2/3$, and $1/3$ for $x=2$.

\subsection{Missing covariate values}\label{subsec:categ_missing}
In practice, it is possible for some of the covariates in the data to be missing. If the $k$-th covariate $x^{(k)}$ in question is categorical, then `missingness' can be treated as an additional factor level. Otherwise, \softc implements left and right splits of the form
\[
\left\{
\left\{x^{(k)} \le \chi \text{ or } x^{(k)}\text{ missing}\right\},
\left\{x^{(k)} > \chi \right\}
\right\}
\quad\text{or}\quad
\left\{
\left\{x^{(k)} \le \chi \right\},
\left\{x^{(k)} > \chi \text{ or } x^{(k)}\text{ missing}\right\}
\right\}.
\]

\subsection{Variable importance}\label{subsec:var_imp}
The variable importance measure for the $k$-th variable (the zeroth one being time $t$) is
\begin{equation}
\mathcal{I}_{k}=\sum_{m=0}^{M-1}\mathcal{I}_{k}(g_{m}) = \sum_{m=0}^{M-1} \sum_{\ell=1}^{L}\Pi_{m,\ell}I(v(m,\ell)=k)\geq0,
\label{eq:vimp}
\end{equation}
where for tree $g_{m}$ with $L$ internal nodes, $\Pi_{m,\ell}$ is the split score \eqref{eq:splitscore} at the $\ell$-th internal node, and $v(m,\ell)$ is the variable used for the partition. Hence the inner sum on the right represents the total reduction in likelihood risk due to splits on the $k$-th variable in the $m$-th tree, and $\mathcal{I}_k$ is the total reduction across the $M$ trees. In other words, the variable importance quantifies a variable's contribution to reducing the likelihood risk. This is more natural than the traditional variable importance measure in \citep{friedman}, which is defined as the reduction in mean squared error between the tree learners and the gradients of the risk at each boosting iteration. To convert $\mathcal{I}_{k}$ into a measure of relative importance between 0 and 1, it is scaled by $\max_{k}\mathcal{I}_{k}$, where a larger value confers higher importance.

\section[Using BoXHED2.0]{Using \pkg{BoXHED2.0}} \label{sec:software}

This section employs a synthetic dataset to walk readers through the use of \softd. A detailed Jupyter notebook tutorial called {\ttfamily BoXHED2\_tutorial.ipynb} is also provided on the GitHub page for \softd. 

\subsection{Structure of training data}\label{sec:inputdata}
Input data on the event histories of study subjects are provided to \softc as a \pandas dataframe \citep{reback2020pandas}. The \pandas dataframe follows the same convention as a Cox analysis of time-dependent covariates:

\begin{itemize}
    \item{\makebox[1.2cm][l]{ID} subject ID.}
    \item{\makebox[1.2cm][l]{$t_{start}$} the start time of an epoch for the subject.}
    \item{\makebox[1.2cm][l]{$t_{end}$} the end time of the epoch.}
    \item{\makebox[1.2cm][l]{$X_i$} value of the $i$-th covariate between $t_{start}$ and $t_{end}$.}
    \item{\makebox[1.2cm][l]{$\delta$}  event label, which is 1 if an event occurred at \tendb; 0 otherwise.}
\end{itemize}

An illustrative example of the entries of the dataframe is:

\begin{Code}
       ID  t_start    t_end       X0  ...    X10  delta
    0   1   0.0100   0.0747   0.2655      0.2059      1
    1   1   0.0747   0.1072   0.7829      0.4380      0
    2   1   0.1072   0.1526   0.7570      0.7789      1
    3   2   0.2066   0.2105   0.9618      0.0859      1
    4   2   0.2345   0.2716   0.3586      0.0242      0
\end{Code}

Each row of the dataframe corresponds to an {\it epoch} of a subject's history. The start and end times of an epoch are given by \tstart and \tendb, and the values of the subject's covariates ({\ttfamily X0} to {\ttfamily X10} in this example) are constant inside an epoch. For each row we must have $0 \le$\tstartb$<$\tendb. Also, epochs cannot overlap. In other words, the beginning of an epoch cannot start earlier than the end of the prior epoch. Any column whose name is not in the set \{\idb, \tstartb, \tendb, \deltab\} is treated as a covariate.

\subsection[Importing BoXHED and defining an instance]{Importing \pkg{BoXHED} and defining an instance}

From the \proglang{Python} package \softe the class \textit{boxhed} can be imported
\begin{Code}
from boxhed.boxhed import boxhed
\end{Code}

and from the imported class, an instance can be defined:
\begin{Code}
boxhed_ = boxhed()
\end{Code}

\subsection{Data preprocessing}

As explained in Section~\ref{subsec:preprocess}, \softc preprocesses the training data to speed up training. This step only needs to be run once per training set. \softc models are fit directly to the preprocessed data.

The user may specify the number of candidate split points ($\le$256) for time and for each non-categorical covariate (argument {\ttfamily num\_quantiles}). The locations of such splits are based on the marginal quantiles of the training data. Alternatively, the user may specify no more than 256 custom candidate split points for time and/or a subset of non-categorical covariates (argument {\ttfamily split\_vals}). For example, if the third line in the code below is uncommented, candidate splits would be limited to four locations on time and three locations on the {\ttfamily X\_2} variable, while the other non-categorical variables will each be endowed with 256 candidate split points:
\begin{Code}
X_post = boxhed_.preprocess(
        data          = train_data, 
        #split_vals   = {"t": [0.2, 0.4, 0.6, 0.8], "X_2": [0, 0.4, 0.9]},
        num_quantiles = 256, 
        nthread       = 20)
\end{Code}

The preprocessor returns a dictionary which contains the preprocessed data, which is used for training and hyperparameter tuning.

\subsection{Hyperparameter tuning}

\softc enables both manual selection of hyperparameters and also hyperparameter tuning through $K$-fold cross-validation. The primary \softe hyperparameters that need to be tuned are:
\begin{itemize}
    \item{\makebox[2.5cm][l]{\maxdepthb} the maximum depth of each boosted tree.}
    \item{\makebox[2.5cm][l]{\nestimatorsb} the number of trees.}
\end{itemize}

The hyperparameter grid provided to cross validation is specified as a dictionary:

\begin{Code}
param_grid = {
    'max_depth':    [1, 2, 3, 4, 5],
    'n_estimators': [50, 100, 150, 200, 250, 300]
}
\end{Code}

This hyperparameter grid amounts to trying trees of depth up to 5 ($2^5$ leaf nodes) and up to 300 trees. The folds are split by subject ID, so that all of the subject's epochs belong to either the training split or the validation split. To perform $K$-fold cross-validation we first import the function \cvb:

\begin{Code}
from boxhed.model_selection import cv
\end{Code}

The \cv function may be called on the preprocessed data. The number $K$ in $K$-fold cross-validation is set as the variable {\ttfamily num\_folds}. The user can specify whether to use GPUs or CPUs in the \cv function. Here we use CPUs by setting {\ttfamily gpu\_list} $=[-1]$. The parameter {\ttfamily nthread} is set to $1$ by default, while setting it to $-1$ corresponds to using all CPU threads. For instructions on how to use GPU, refer to the tutorial on GitHub.

The user can specify the argument {\ttfamily seed} to fix the seed of the random number generator used to produce the cross validation splits. Here, we choose a value of 6 for replication purposes:

\begin{Code}
cv_rslts = cv(param_grid, 
              X_post,
              num_folds = 5,
              seed      = 6,
              ID        = ID,
              gpu_list  = [-1],
              nthread   = -1)
\end{Code}

The function \cv returns a dictionary containing the possible hyperparameter combinations along with their $K$-fold means and standard errors of log-likelihood values. They are all in form of \numpy vectors \citep{harris2020array}. The means can be inspected as follows:

\begin{Code}
import numpy as np
nrow, ncol = len(param_grid['max_depth']), len(param_grid['n_estimators'])
print( np.around( cv_rslts['score_mean'].reshape(nrow, ncol), 2) )
\end{Code}

\begin{Code}
    [[ -518.84 -500.75 -496.83 -495.81 -495.93 -495.97]
     [ -499.32 -498.62 -500.91 -502.83 -505.1  -507.48]
     [ -500.69 -508.07 -513.77 -520.41 -526.88 -533.33]
     [ -507.09 -518.29 -531.18 -545.01 -555.32 -569.18]
     [ -516.2  -533.55 -555.09 -572.67 -593.68 -614.24]]
\end{Code}

For standard errors:
\begin{Code}
print( np.around( cv_rslts['score_ste'].reshape(nrow, ncol), 2) )
\end{Code}

\begin{Code}
   [[ 5.72 5.05 4.86 4.72 4.77 4.73]
    [ 4.7  4.91 4.99 5.24 5.49 5.38]
    [ 4.73 5.   5.04 4.91 4.73 4.5 ]
    [ 5.73 6.15 5.92 6.03 5.69 6.48]
    [ 6.37 5.89 6.06 7.07 7.   7.51]]
\end{Code}

The numbers above can be expressed in the tabular form (standard errors in parentheses):
\scriptsize
\begin{Code}
n_estimators     50             100             150             200             250             300
max_depth     
1     -518.84(5.72)   -500.75(5.05)   -496.83(4.86)   -495.81(4.72)   -495.93(4.77)   -495.97(4.73)
2     -499.32(4.70)   -498.62(4.91)   -500.91(4.99)   -502.83(5.24)   -505.10(5.49)   -507.48(5.38)
3     -500.69(4.73)   -508.07(5.00)   -513.77(5.04)   -520.41(4.91)   -526.88(4.73)   -533.33(4.50)
4     -507.09(5.73)   -518.29(6.15)   -531.18(5.92)   -545.01(6.03)   -555.32(5.69)   -569.18(6.48) 
5     -516.20(6.37)   -533.55(5.89)   -555.09(6.06)   -572.67(7.07)   -593.68(7.00)   -614.24(7.51) 
\end{Code}
\normalsize

The mean log-likelihood is maximized at \{\maxdepthb=1, \nestimatorsb=200\} with value $-495.81$. However, the \textit{one-standard-error rule} ($\S$7.10 in \citep{hastie2009elements}) suggests choosing the `simplest model' whose mean log-likelihood is no less than 1 standard error of the maximum. In this example, this means choosing the simplest model whose mean log-likelihood is no less than $-495.81-4.72 = -500.53$. The hyperparameters with mean log-likelihood larger than this are:

\scriptsize
\begin{Code}
n_estimators     50              100             150             200             250             300
max_depth 
1                                            -496.83         -495.81         -495.93         -495.97
2           -499.32          -498.62
3
4               
5  
\end{Code}
\normalsize
\color{black}

There is no generally accepted way to compare the complexities of these choices against one another. One heuristic is to define the complexity of a hyperparameter combination as
$$\log_2\left( \text{n\_estimators}\times 2^\text{max\_depth} \right) = \log_2( \text{n\_estimators}) + \text{max\_depth},$$
which is $\log_2$(total number of leaf nodes in tree ensemble). Under this criterion, \{\maxdepthb=2, \nestimatorsb=50\} has the lowest complexity, and is hence the most parsimonious choice. However, if we only consider hyperparameters with \maxdepthb $\le$1 and \nestimatorsb$\le$200, then \{\maxdepthb=1, \nestimatorsb=150\} is the most parsimonious. This restriction ensures that the chosen combination is no less parsimonious than the log-likelihood maximizer under any sound definition of complexity.

The function {\ttfamily best\_param\_1se\_rule()} automates the described process of finding the most parsimonious combination. It first needs to be imported:
\begin{Code}
from boxhed.model_selection import best_param_1se_rule
\end{Code}

The user needs to then supply a model complexity measure:

\begin{Code}
def model_complexity(max_depth, n_estimators):
    from math import log2
    return log2(n_estimators) + max_depth
\end{Code}

And finally run the {\ttfamily best\_param\_1se\_rule()} function:

\begin{Code}
best_params, _ = best_param_1se_rule(cv_rslts,
                                     model_complexity,
                                     bounded_search = True)
\end{Code}

Setting {\ttfamily bounded\_search}$=${\ttfamily True} forces the search to only consider hyperparameter combinations satisfying \maxdepthb $\le$1 and \nestimatorsb$\le$200 (the hyperparameter combination maximizing the log-likelihood).

Having found the best hyperparameter combination, the \softe instance can be set to this hyperparameter combination:

\begin{Code}
boxhed_.set_params(**best_params)
\end{Code}

Alternatively, the user can manually supply the hyperparameter combination in the following way:

\begin{Code}
boxhed_.set_params(**{'max_depth':1, 'n_estimators':150})
\end{Code}

\subsection[Fitting BoXHED]{Fitting \pkg{BoXHED}}

\softe can be fit using multicore CPUs or GPUs. GPUs are normally accessed through an integer identifier which is in the range \{0, 1, \dots, \# GPUs -1\}. To use a GPU, you may specify its number. For example, to use the first GPU:
\begin{Code}
boxhed_.set_params(**{'gpu_id'=0})
\end{Code}

The default value for the parameter {\ttfamily gpu\_id} is $-1$ which corresponds to using CPUs only. When using CPUs, the number of CPU threads can be set by:

\begin{Code}
boxhed_.set_params(**{'nthread'=20})
\end{Code}

Following the same convention as the function \cvb, setting the parameter {\ttfamily nthread} to $-1$ corresponds to using all CPU threads. Finally, to fit \softf, execute the following to pass the preprocessed data {\ttfamily X\_post} to the {\ttfamily fit()} function:

\begin{Code}
boxhed_.fit(X_post['X'], X_post['delta'], X_post['w'])
\end{Code}

\softe stores the variable importance measures, as defined in Section~\ref{subsec:var_imp}, in a class variable {\ttfamily VarImps}. It is a dictionary of variables with their corresponding importances. It can be printed:

\begin{Code}
print (boxhed_.VarImps)
\end{Code}
\begin{Code}
    { 'X_0':  1971.2105061100006, 
      'time': 1558.9728342899996, 
      'X_4':  21.197394600000003, 
      'X_9':  9.797206169999999, 
      'X_1':  6.96407747, 
      'X_10': 6.74315118, 
      'X_6':  2.85510993, 
      'X_7':  5.40960407 }
\end{Code}

The set of time values where tree splits are made can be accessed by:
\begin{Code}
print (boxhed_.time_splits)
\end{Code}
\begin{Code}
    [0.02809964 0.03668491 0.04400259 0.05081875 0.07196306 0.08450372
     0.10639625 0.11365116 0.14458408 0.15542936 0.19222417 0.20236476
     0.21580724 0.23802629 0.3294313  0.36373213 0.42403865 0.69717562
     0.70240098 0.72262114 0.82380444 0.83513868 0.84786463 0.89849269
     0.93779886 0.97168624]
\end{Code}

\subsection{Hazard estimation}
We can extract the value of the estimated hazard function $\hat\lambda(t,x_0,x_1,\cdots)$ evaluated at a given time $t$ with covariate values $X(t)=(x_0,x_1,\cdots)$. For example, given the following \pandas dataframe {\ttfamily test\_X}:

\begin{Code}
        t          X_0  ...  X_10
    0   0.000000   0.0       0.508183
    1   0.010101   0.0       0.414983
    2   0.020202   0.0       0.407774
    3   0.030303   0.0       0.589770
    4   0.040404   0.0       0.840509
\end{Code}

we can retrieve the estimated value $\hat\lambda(t,x_0,\cdots,x_{10})$ for each row of {\ttfamily test\_X} using the {\ttfamily hazard()} function:

\begin{Code}
haz_vals = boxhed_.hazard(test_X)
\end{Code}

Plotting the estimated hazard values as a function of time and one of the covariates, $X_0$, yields the following plot:

\begin{figure}[H]
\centering
\includegraphics[width=0.8\textwidth]{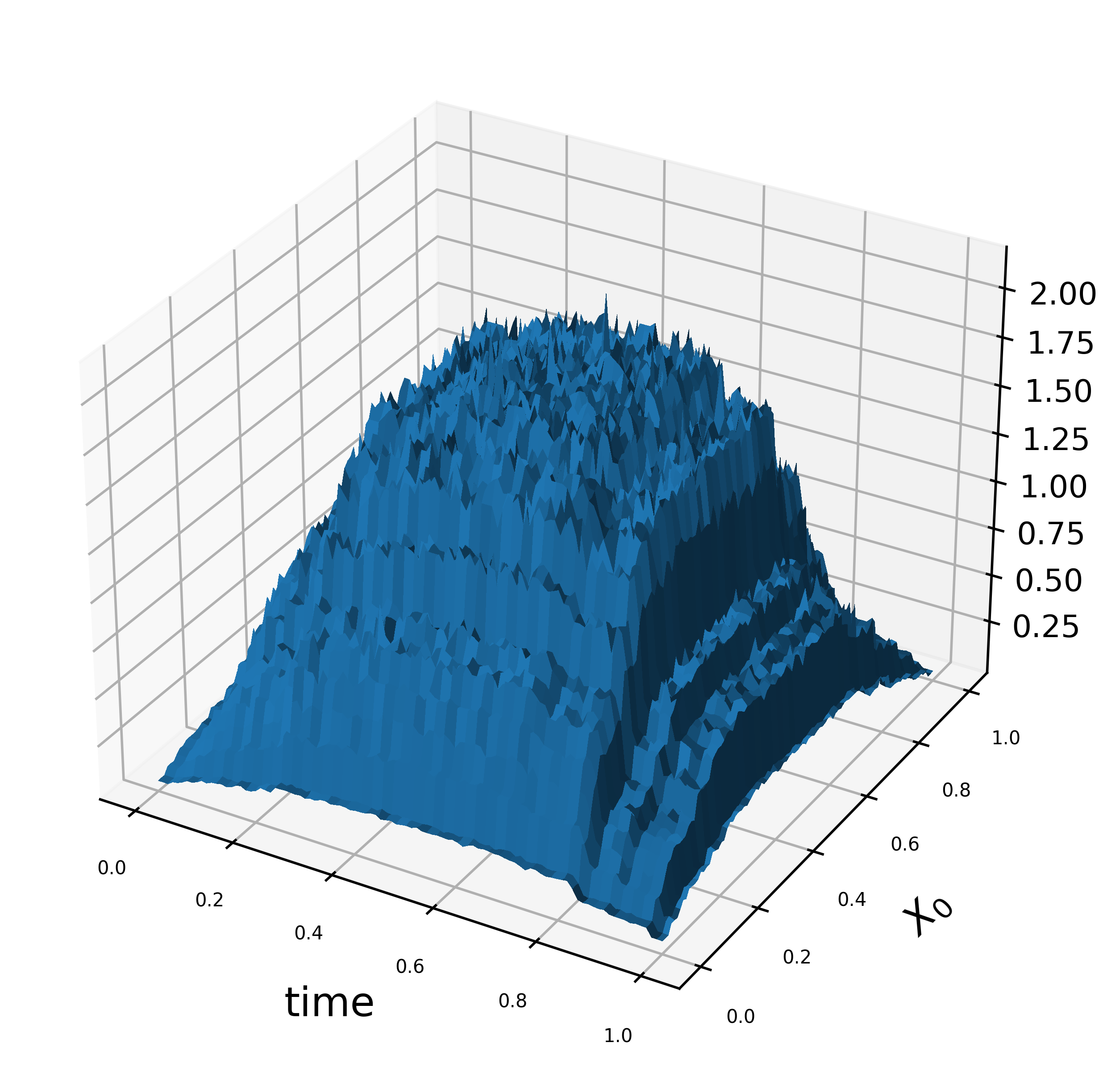}
\caption{The estimated hazard values as a function of time and one of the covariates, $X_0$.}
\end{figure}
 
\subsection{Survivor curve estimation for time-static covariates}

Recall from Section~\ref{sec:intro} that the survivor curve $S(t|X)=\mathbb{P}(T>t|X)$ is not meaningful when the covariates $X(t)$ change over time. However, for time-static covariates $X(t)=X$, $S(t|X)$ can be obtained from the estimated hazard function.

As an example, suppose we want to obtain the survivor curve conditional on the covariate vector $(X_0=0.20202,\cdots,X_{10}=0.128217)$. We wish to evaluate the curve at $t=0,0.01,0.02,\cdots,0.99$. We use the code below to generate the \pandas dataframe:
\begin{Code}
t            = [t/100 for t in range(0, 100)]
df_surv      = pd.concat([test_X.loc[2000].to_frame().T]*len(t))
df_surv      = df_surv.reset_index(drop=True)
df_surv['t'] = t
\end{Code}

\begin{Code}
        t     X_0      ...  X_10
    0   0.00  0.20202       0.128217
    1   0.01  0.20202       0.128217
    2   0.02  0.20202       0.128217
    ..   ...      ...       ...
    98  0.98  0.20202       0.128217
    99  0.99  0.20202       0.128217
\end{Code}

To estimate the value of the survivor curve for each row, run:

\begin{Code}
    surv_vals = boxhed_.survivor(df_surv)
\end{Code}

\begin{figure}[H]
\centering
\includegraphics[width=0.6\textwidth]{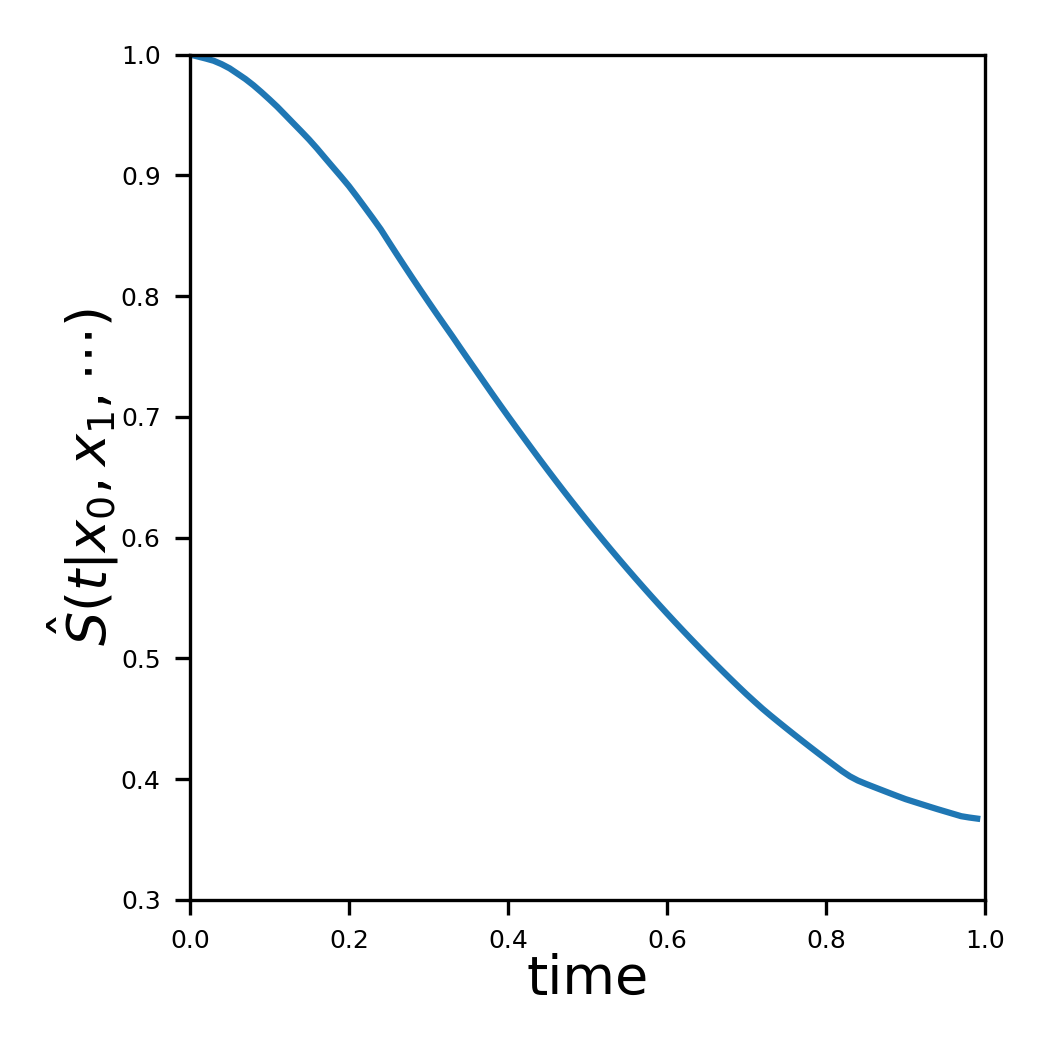}
\caption{The estimated survivor curve as a function of time.}
\end{figure}

\subsection[Saving and loading the fitted BoXHED model]{Saving and loading the fitted \pkg{BoXHED} model}
Use {\ttfamily dump\_model()} to save the fitted \softe instance to disk, and use {\ttfamily load\_model()} to retrieve it.

Saving a \softe model:
\begin{Code}
    boxhed_.dump_model("./boxhed.pkl")
\end{Code}

Retrieving a \softe model:
\begin{Code}
    boxhed_2 = boxhed()
    boxhed_2.load_model("./boxhed.pkl")
\end{Code}

\newpage
\section[Scalability analysis]{Scalability analysis}\label{sec:computational_complexity}

To benchmark how \softc scales with data, we perform runtime comparisons in three settings: 1) Comparing \softc (CPU and GPU) against \softa as the number of rows in the dataset is increased; 2) Comparing \softc against the boosted parametric survival models in \citet{hothorn2010model} as the number of rows in the dataset is increased; and 3) Assessing how \softc scales with the number of covariates in the model.

To create a synthetic dataset of a given size, we used one of the experiments in \citet{wang2020boxhed}. The dataset consists of 41 covariates and up to 10 million rows, details are provided in the Appendix. Computations were performed on a server with two Intel Xeon CPU E5-2650 v4 2.20GHz processors with 512 GB of RAM, and a GeForce GTX 1080 Ti GPU with 11GB of RAM. Raw quantiles were used to select $256$ splits. 20 CPU threads were utilized for \softc CPU.

\begin{figure}[H]
\centering     
\includegraphics[width=0.6\columnwidth]{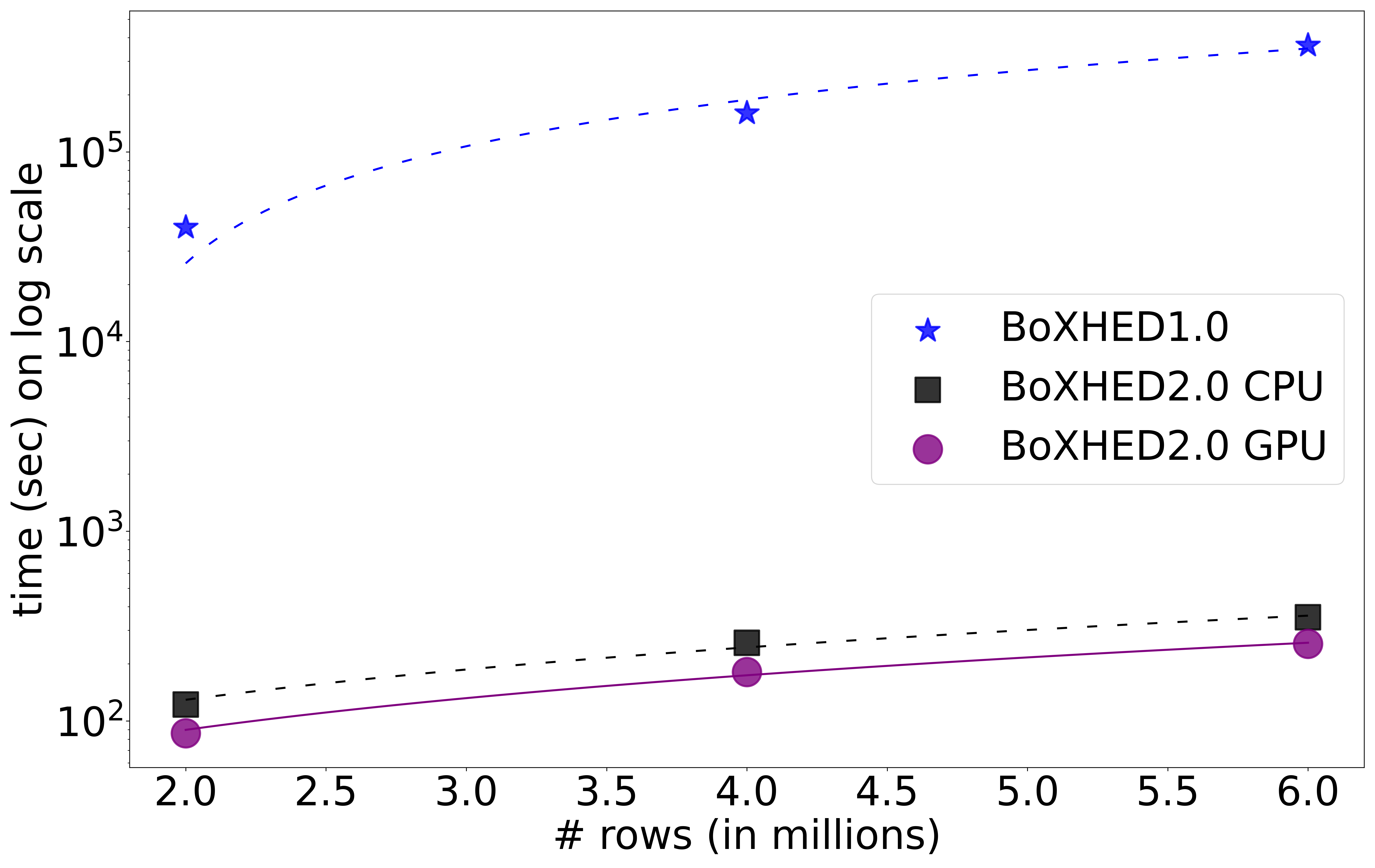}
\caption{Runtimes of \softc vs. \softa as the number of data rows increase.}
\label{fig:2vs1}
\end{figure}

\paragraph{\softc vs. \softb: Scaling the number of data rows.} Figure~\ref{fig:2vs1} depicts how \softc and \softa scale as the number of data rows is increased. The hyperparameters chosen for \softe are \{\maxdepthb=1, \nestimatorsb=250\}. On the other hand, the hyperparameters for \softa are set as \{\maxdepthb=1, \nestimatorsb=1\} and each covariate (including time) are given 10 candidate splits. In other words, the \softa runtimes measure how long it takes \softa to decide on the first split from among 10 candidate splits rather than from 256. Even after giving \softa such a substantial leg up, \softc is still on average 900 times faster.

\begin{figure}[H]
\centering     
\includegraphics[width=0.6\columnwidth]{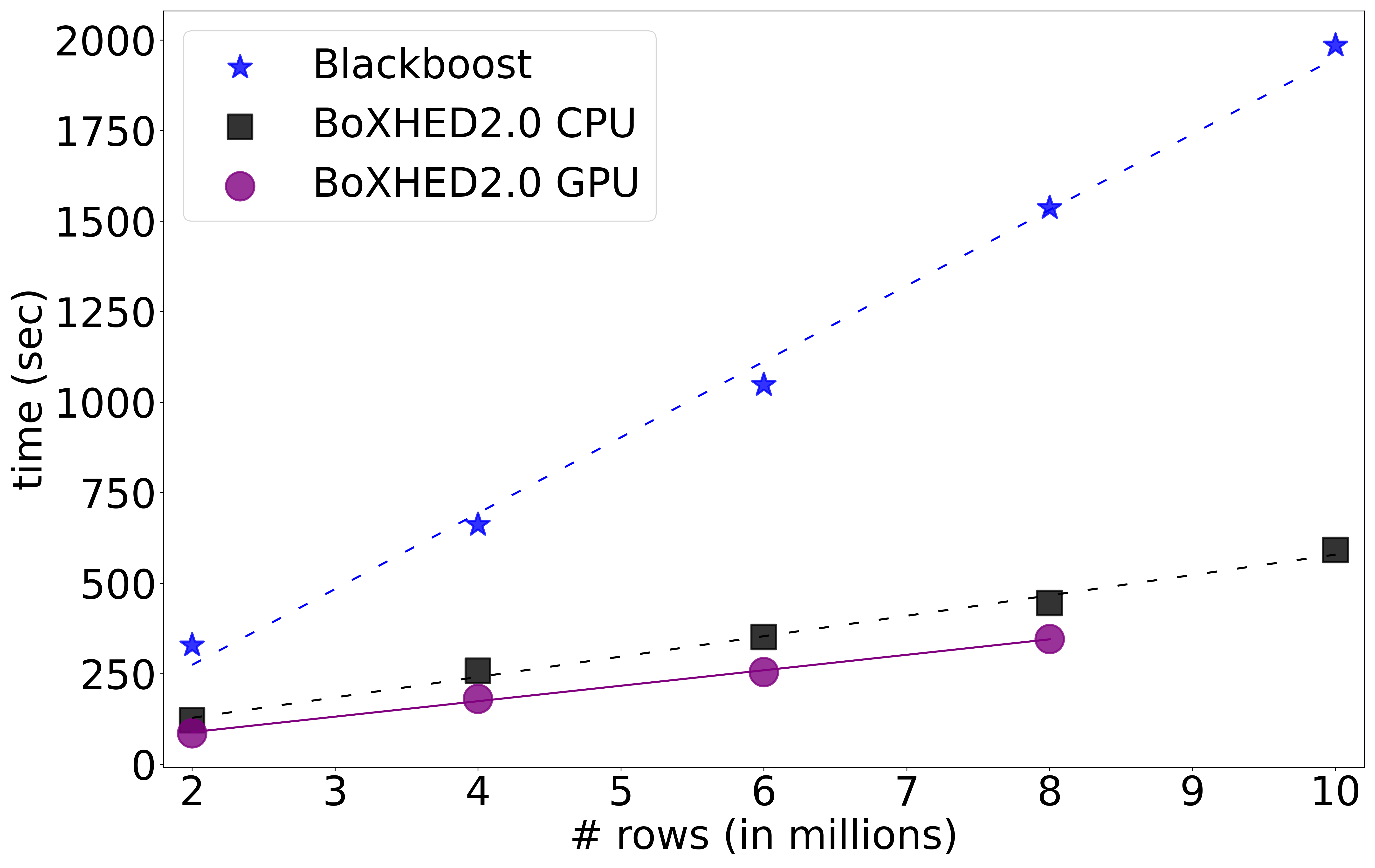}
\caption{Scalability analysis of \softc vs. the boosted parametric survival model \blackboost (log-normal family).}
\label{fig:2vsblackboost}
\end{figure}

\paragraph{\softc vs. boosted parametric survival models: Scaling the number of data rows.} 
Figure~\ref{fig:2vsblackboost} compares \softc run times against those for the boosted parametric survival model \blackboost in \proglang{R} \citep{hothorn2010model}. The hyperparameters chosen for all estimators are \{\maxdepthb=1, \nestimatorsb=250\}. For \softc GPU, only the results for up to 8 million rows of data are included, as larger datasets did not fit into the memory of the GPU we used for testing. We see from the figure that all methods scale linearly with the number of rows. Remarkably though, the nonparametric \softd's speed exceeds that of the boosted parametric model by a large margin. This is even more remarkable given that \blackboost only applies to time-static covariates, so it does not have to contend with the extra computations required for time-dependent covariates.

\begin{figure}[H]
\centering     
\includegraphics[width=0.6\columnwidth]{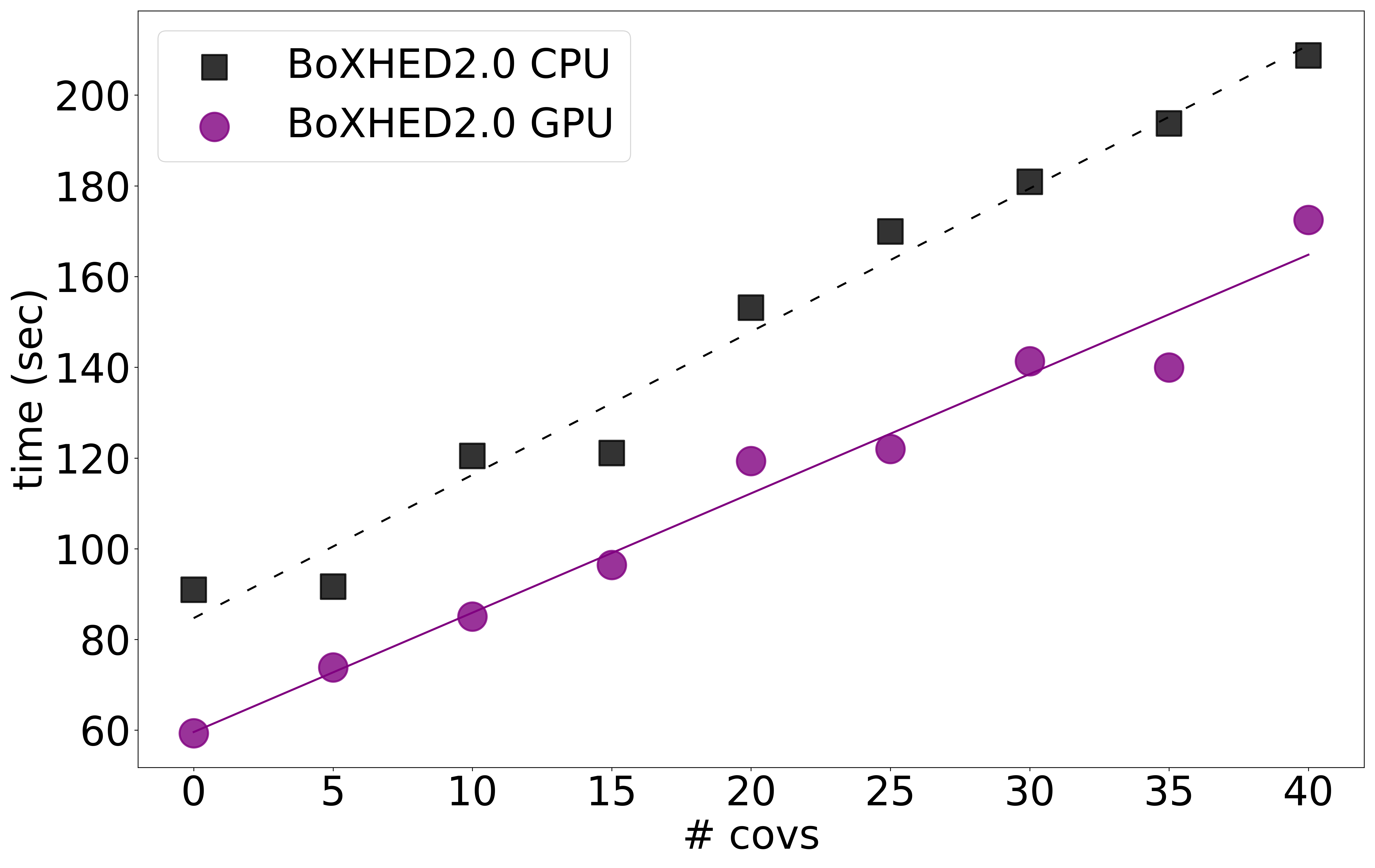}
\caption{Scalability analysis of \softc for varying number of covariates.}
\label{fig:scalability_by_ncovs}
\end{figure}

\paragraph{\softd: Scaling the number of covariates.} Figure~\ref{fig:scalability_by_ncovs} reveals how the runtime of \softc scales with the number of covariates. The number of rows in the dataset is fixed at 4 million rows. The chosen hyperparameters are \{\maxdepthb=1, \nestimatorsb=250\}. We see that \softc runtimes scale linearly with the number of covariates, with \softc GPU being around 20\% faster than \softc CPU.

\section[BoXHED2.0 in Practice]{\pkg{BoXHED2.0} as a realtime risk monitor}\label{sec:boxhed_in_practice}

We demonstrate \softc in practice by assessing the realtime risk of experiencing invasive-ventilation events in the intensive care unit (ICU). This involves survival data that go beyond the right-censoring setting prevalent in survival analysis: Over the course of an ICU admission, a patient may experience multiple episodes of invasive ventilation (iV), which is when a tube is inserted into the trachea to assist with breathing. The onset of iV thus constitutes a recurrent event. Moreover, while undergoing an episode of iV, the patient by definition cannot be at-risk of experiencing a concurrent episode of iV, so the at-risk indicator $Y(t)$ must be zero during an episode.

Patients undergoing iV have increased risk of severe adverse events and mortality, with the importance of proper and timely ventilation being punctuated during the COVID-19 pandemic \citep{yu2021machine}. Thus having a realtime risk measure for the onset of iV may facilitate early intervention, potentially via non-invasive means. To our knowledge there is no other nonparametric machine learning solution besides \softc for handling this type of survival data. We demonstrate this application on the MIMIC-IV dataset (Medical Information Mart for Intensive Care, version IV, \citep{Johnson2021-si}), identifying informative covariates and providing real-time risk monitoring. For details on the data extraction process (cohort and variables) see the Appendix.

\begin{figure}[H]
\centering
\includegraphics[width=0.65\textwidth]{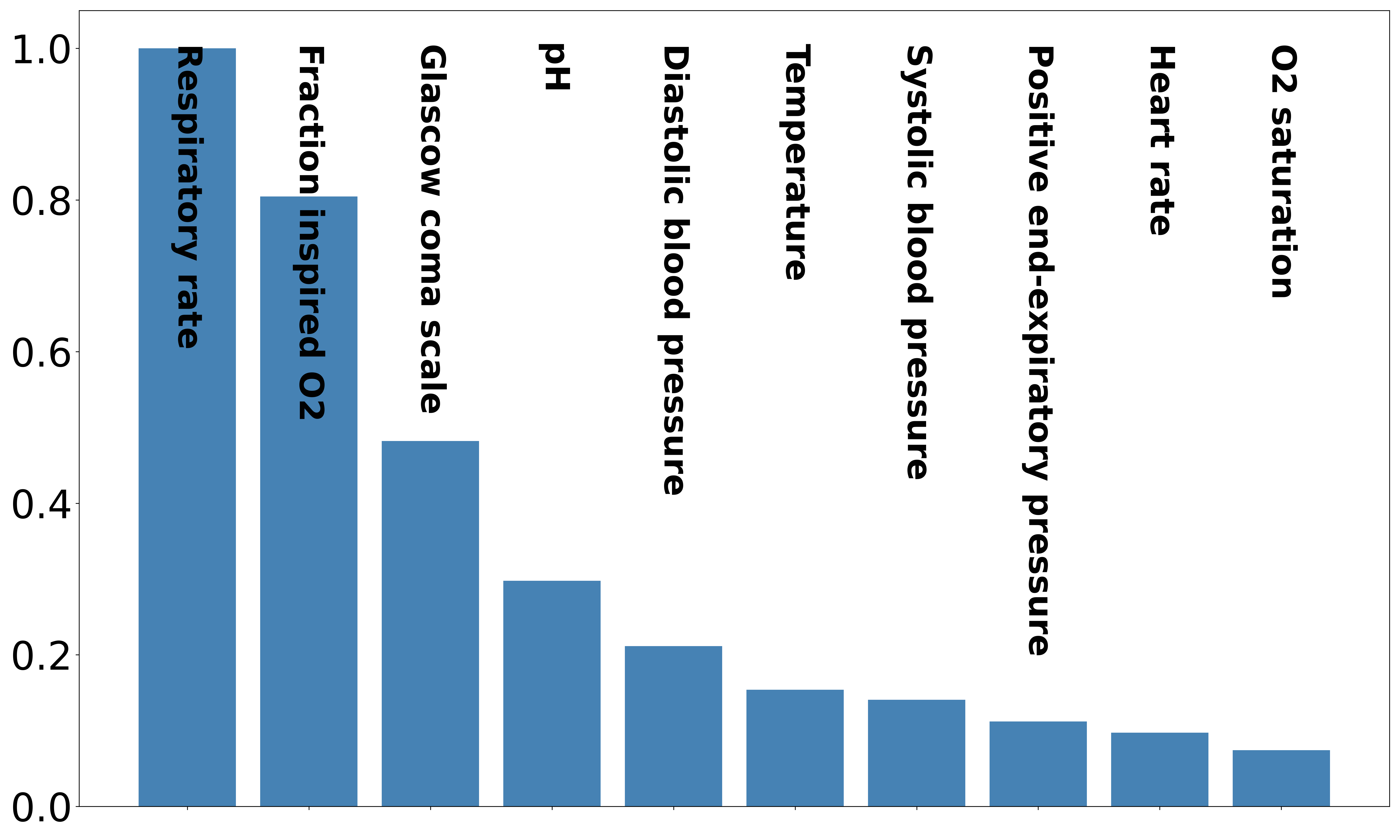}
\caption{Relative variable importances for the onset of iV.\label{fig:relvarimp}}
\end{figure}

After using \softc to estimate the hazard rate for iV, the ten most informative covariates, along with their relative importances, are displayed in Figure~\ref{fig:relvarimp}.

\begin{figure}[H]
\centering
\includegraphics[width=0.65\textwidth]{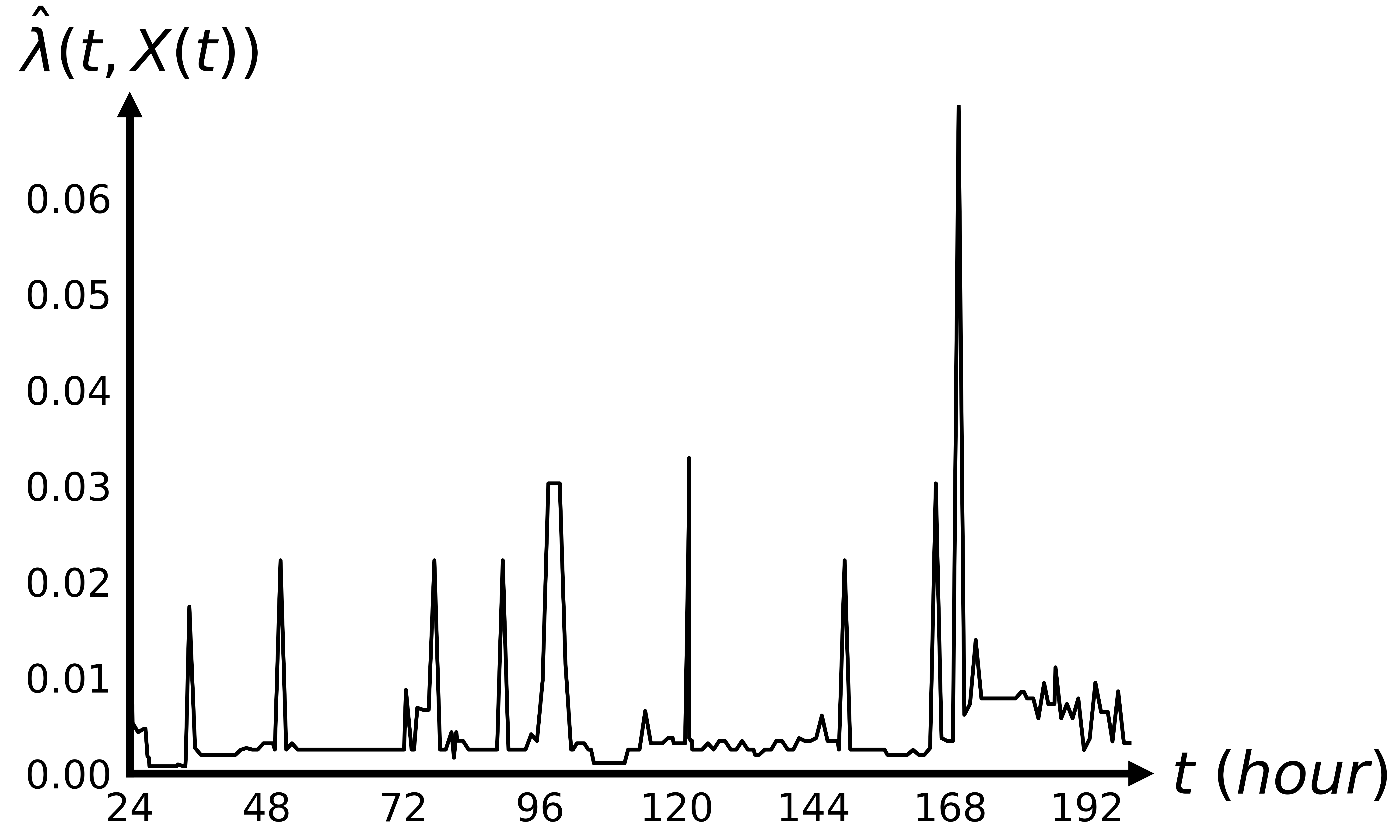}
\caption{\softc can be used as a real-time risk indicator.\label{fig:realtimerisk}}
\end{figure}

As expected, the most important features represent the patient's current (and degrading) respiratory state. The estimated hazard values serve as a realtime risk indicator. As a patient's ICU stay progresses, their vitals evolve and hence so does the patient's risk. An example of how the iV risk evolves for a randomly selected patient is shown in Figure~\ref{fig:realtimerisk}.

\section{Discussion}\label{sec:discussion}

With its core written in \proglang{C++}, \softc is a \proglang{Python} package which is the only nonparametric boosting implementation for time-dependent survival settings beyond the classic right-censoring setup. It is based on the theoretically justified procedure in \citet{lee2021boosted}, and runs at speeds that match even those of boosted parametric models. \softc supports the use of GPUs and multicore CPUs, and is available from \href{http://github.com/BoXHED}{github.com/BoXHED}.


\section{Acknowledgments}
This work was supported in part by NIH award 1R21EB028486-01.

\bibliography{refs}


\newpage

\begin{appendix}

\section{Data preprocessing}\label{sec:preprocessing}
Following the standard treatment of time-dependent survival data in the Cox model, the functional data sample $\{X_i(t),Y_i(t), N_i(t)\}_t$ for subject $i$ is fed into \softc in the tabular form
\begin{equation}\label{eq:tabulardata}
\begin{pmatrix}
	\underline\tau_{i1} & \overline\tau_{i1} & \chi_{i1} & \Delta_{i1} \\
	\underline\tau_{i2} & \overline\tau_{i2} & \chi_{i2} & \Delta_{i2} \\
	\multicolumn{4}{c}{\vdots} \\
	\underline\tau_{i{J_i}} & \overline\tau_{i{J_i}} & \chi_{i{J_i}} & \Delta_{i{J_i}}
\end{pmatrix},
\end{equation}
where the $j$-th epoch (row) represents a time interval $(\underline\tau_{ij}, \overline\tau_{ij}]$ over which the subject is at-risk, i.e., $Y_i(t)=1$. Note that the first epoch need not start at $\underline\tau_{i1}=0$, and the endpoint $\overline\tau_{ij}$ of the $j$-th epoch need not equal the beginning $\underline\tau_{i,j+1}$ of the $(j+1)$-th epoch. The value of the covariate in the $j$-th epoch is $\chi_{ij}$, and $\Delta_{ij}=1$ if the subject experienced an event at $\overline\tau_{ij}$, and is zero otherwise. The goal is to preprocess the data so that the quantities \eqref{eq:UV},
\[
U_k = \sum_{i=1}^{n}\int_{(t,X_i(t))\in A_k} Y_i(t) e^{F_{m}(t,X_{i}(t))}dt,
\qquad
V_k = \sum_{i=1}^n \int_{(t,X_i(t))\in A_k} dN_i(t),
\]
defined for the left and right daughter regions $A_L$ and $A_R$ can be computed efficiently without numerical integration.

The insight behind the preprocessing step stems from the observation that the tree learners $g_0(t,x),g_1(t,x),\cdots$ in $F_m(t,x) = F_0 - \nu\sum_{q=0}^{m-1} g_{q}(t,x)$ are piecewise-constant over the disjoint time-covariate regions \eqref{eq:regions}:
\[
B_\ell=\left\{(t,x) : 
\begin{pmatrix}
t^{(\ell_0)}<t\leq t^{(\ell_0+1)}\\
x^{(1,\ell_1)}<x^{(1)}\leq x^{(1,\ell_1+1)}\\
\vdots\\
x^{(p,\ell_p)}<x^{(p)}\leq x^{(p,\ell_p+1)}
\end{pmatrix}\right\},
\]
where $\{t^{(\ell_0)}\}_{\ell_0}$ is the set of candidate split points for time $t$, and $\{x^{(k,\ell_k)}\}_{\ell_k}$ is the set of candidate splits for the $k$-th covariate $x^{(k)}$. It follows that for a fixed value of $x$, $F_m(t,x)$ is a constant function of time between any two consecutive candidate split points $t^{(\ell_0)}$ and $t^{(\ell_0+1)}$. Suppose for a moment that each epoch $(\underline\tau_{ij},\overline\tau_{ij}]$ in the input data is completely contained within some interval $(t^{(\ell_0)}, t^{(\ell_0+1)}]$. Since each daughter region $A_{k\in\{L,R\}}$ is the union of some subset of the regions \eqref{eq:regions}, $U_k$ and $V_k$ reduce to the weighted sums
\begin{equation}\label{eq:UVsum}
U_k = \sum_{(\underline\tau_{ij}+,\chi_{ij})\in A_k} w_{ij} e^{F_{m}(\underline\tau_{ij}+,\chi_{ij})},
\qquad
V_k = \sum_{(\underline\tau_{ij}+,\chi_{ij})\in A_k} \Delta_{ij},
\end{equation}
where $w_{ij}=\overline\tau_{ij}-\underline\tau_{ij}$.
The choice of $\underline\tau_{ij}+$ in \eqref{eq:UVsum} is due to the fact that $F_m(t,x)$ is constant over half-open intervals of the form $(t^{(\ell_0)}, t^{(\ell_0+1)}]$.

Similarly, the likelihood risk \eqref{eq:loglik} evaluated at $F_m$, $R_n(F_m)$, can be computed as the sum
\begin{equation}\label{eq:logliksum}
\frac{1}{n}\sum_{ij} \biggl\{ w_{ij} e^{F_{m}(\underline\tau_{ij}+,\chi_{ij})} - \Delta_{ij} F_{m}(\underline\tau_{ij}+,\chi_{ij}) \biggr\}.
\end{equation}
Thus the `sufficient statistics' for \softc are $\{\underline\tau_{ij}, w_{ij}, \chi_{ij}, \Delta_{ij}\}_{ij}$, and \eqref{eq:UVsum}-\eqref{eq:logliksum} can be easily computed from the tree ensemble predictions for  $F_m(t,x)$.

To excise explicit numerical integration from \softd, we need to put the data \eqref{eq:tabulardata} in a form where each epoch is completely contained within some interval $(t^{(\ell_0)}, t^{(\ell_0+1)}]$. This is done by applying four operations to the rows of \eqref{eq:tabulardata}. The first two are:
\begin{enumerate}
	\item \label{step:break} Any epoch $(\underline\tau_{ij},\overline\tau_{ij}]$ that contains one of the candidate split points $t^{(\ell_0)}$ is split into $(\underline\tau_{ij},t^{(\ell_0)}]$ and $(t^{(\ell_0)},\overline\tau_{ij}]$. If an epoch spans multiple split points then it is split into a number of shorter epochs.
	\item \label{step:collapse} Transform the (newly created) rows $\{\underline\tau_{ij}, \overline\tau_{ij}, \chi_{ij}, \Delta_{ij}\}_{ij}$ into $\{\underline\tau_{ij}, w_{ij}, \chi_{ij}, \Delta_{ij}\}_{ij}$.
\end{enumerate}
The time complexity for achieving the above is $\mathcal{O}(n\log|\{t^{(\ell_0)}\}_{\ell_0}|)$. Figure~\ref{fig:initialsteps} illustrates the operations on a simple numerical example. The left table describes the event histories for two subjects. The first subject experienced an event at the end of their first epoch ($t=0.13$), returns to the sample at $t=0.15$, and becomes lost to follow up at $t=0.25$. A similar story can be told for the second subject.

The second set of required operations stem from the fact that most boosting implementations, including \xgbb, use tree splits of the form `$<$' for the left daughter and `$\ge$' for the right. In particular, this implies time splits of the form $[\cdot, \cdot)$. This is incompatible with the epochs for survival data, which are intrinsically of the form $(\cdot,\cdot]$. To see why, consider for example a subject who experiences an event at a candidate split point $t^{(\ell_0)}$. Since the candidate split points are not sampled from an absolutely continuous distribution, but are instead chosen from the observed data, the probability of this happening is not zero. Under the splitting convention $[\cdot, \cdot)$, the observed event will be counted towards some region $[t^{(\ell_0)}, t^{(\ell_0')})$, even though the subject was never at-risk there. The hazard MLE for that region would then be one (number of events in region) divided by zero (time spent at-risk in region).\footnote{The format for the \softa data allowed for similar situations where $V_k>0$ while $U_k=0$. An ad-hoc imputation was used to remove the pathology (\S 4.1.2 in Wang et al., 2020). The \softc data format is designed to rule out this possibility from the start.} For mathematical consistency, the lower end of the time interval must be open and the upper end closed. For notational consistency we will also apply the same convention to each covariate.

Fortunately, through additional processing, we can still use \xgbb's splitting convention to fit trees with leaf nodes of the form $(t^{(\ell_0)},t^{(\ell_0')}] \times (x^{(1,\ell_1)},x^{(1,\ell_1')}] \times \cdots \times (x^{(p,\ell_p)},x^{(p,\ell_p')}]$. Let us first discuss the processing of covariates. Suppose that a leaf region $B_{m,\ell}$ contains an epoch whose $k$-th covariate value $\chi_{ij}^{(k)}$ coincides with one of the candidate split points $x^{(k,\ell_k)}$. A \xgb cut at that location would assign the epoch to the right daughter $B_{m,\ell}\cap\{x^{(k)} \ge x^{(k,\ell_k)}\}$. To ensure that it will be assigned to the left daughter $B_{m,\ell}\cap\{x^{(k)} < x^{(k,\ell_k)}\}$ instead, we apply the map
\begin{enumerate}
	\setcounter{enumi}{2}
	\item \label{step:xmap} $\chi_{ij}^{(k)} \mapsto x^{(k,\ell_k-1)}$ for $\chi_{ij}^{(k)}\in(x^{(k,\ell_k-1)}, x^{(k,\ell_k)}]$.
\end{enumerate}
Note that the covariate values in the interior of $(x^{(k,\ell_k-1)}, x^{(k,\ell_k)}]$ are also mapped to the candidate split point $x^{(k,\ell_k-1)}$. This caps the number of unique covariate values in the data to the maximum number of candidate split points set by the user. If fewer than 256 candidates are used, the covariate values can be stored as a byte, which is useful for GPU training, given memory scarcity.

With the application of Step \ref{step:xmap}, the fitted value for a region $[x^{(k,\ell_k)},x^{(k,\ell_k')})$ in fact represents the fitted value for $(x^{(k,\ell_k)},x^{(k,\ell_k')}]$ in the \softc hazard estimator. Accordingly, the {\ttfamily hazard } function in \softc maps any new covariate value that coincides with a candidate split point $x^{(k,\ell_k)}$ back to $x^{(k,\ell_k-1)}$, before running it through the fitted trees to obtain the estimated hazard value.

For the processing of time, this was already partly accomplished by our indexing of the epoch $(\underline\tau_{ij}, \overline\tau_{ij}]$ with $\underline\tau_{ij}$ rather than $\overline\tau_{ij}$ in \eqref{eq:UVsum} and \eqref{eq:logliksum}. Therefore, if a leaf region $B_{m,\ell}$ is split on time into daughters $B_{m,\ell}\cap\{t<\underline\tau_{ij}\}$ and $B_{m,\ell}\cap\{t\ge\underline\tau_{ij}\}$, the epoch will be correctly assigned to the latter. Thus even if $\underline\tau_{ij}$ coincides with a candidate split point $t^{(\ell_0)}$, there is no need to re-map. However, given the GPU memory discussion above, we still need to perform the following:
\begin{enumerate}
	\setcounter{enumi}{3}
	\item \label{step:tmap} $\underline\tau_{ij} \mapsto t^{(\ell_0-1)}$ for $\underline\tau_{ij} \in (t^{(\ell_0-1)}, t^{(\ell_0)})$.
\end{enumerate}
Steps \ref{step:xmap} and \ref{step:tmap} are illustrated in Figure~\ref{fig:lattersteps}.

Similar to the covariate regions, the fitted value for $[t^{(\ell_0)},t^{(\ell_0')})$ represents the estimated hazard value for $(t^{(\ell_0)},t^{(\ell_0')}]$ in the \softc estimator. Hence the {\ttfamily hazard} function in \softc also maps any new time value that coincides with a candidate split point $t^{(\ell_0)}$ back to $t^{(\ell_0-1)}$.
\begin{figure}[H]
	\caption{Numerical illustration of Steps \ref{step:break} and \ref{step:collapse}. The candidate split points $\{t^{(\ell_0)}\}_{\ell_0}$ are set as \{0, 0.10, 0.15\}, where $0$ is always included by default (but will not be split on). The first epoch in the original data (left) is split into the first two rows in the middle table, and the last epoch in the original data is split into the last two rows in the middle table.}
	\setstacktabbedgap{0.5pt}
	\label{fig:initialsteps}
	\vskip 0.1in
	
	\begin{center}
		\begin{small}
			$\begin{pmatrix}
				i & \underline\tau & \overline\tau & \chi & \Delta \\
				\hline\\
				1 & 0.01 & 0.13 & 0.27 & 1 \\
				1 & 0.15 & 0.25 & 0.51 & 0\\
				2 & 0.06 & 0.10 & 0.81 & 1\\
				2 & 0.13 & 0.25 & 0.92 & 0
			\end{pmatrix}
			\text{$\rightarrow$}
			\begin{pmatrix}
				i & \underline\tau & \overline\tau & \chi & \Delta \\
				\hline\\
				1 & 0.01 & {\bf 0.10} & 0.27 & {\bf 0}\\
				{\bf 1} & {\bf 0.10} & {\bf 0.13} & {\bf 0.27} & {\bf 1}\\
				1 & 0.15 & 0.25 & 0.51 & 0\\
				2 & 0.06 & 0.10 & 0.81 & 1\\
				2 & 0.13 & {\bf 0.15} & 0.92 & 0\\
				{\bf 2} & {\bf 0.15} & {\bf 0.25} & {\bf 0.92} & {\bf 0}
			\end{pmatrix}
			\text{$\rightarrow$}
			\begin{pmatrix}
				i & \underline\tau & w & \chi & \Delta \\
				\hline\\
				1 & 0.01 & {\bf 0.09} & 0.27 & 0\\
				1 & 0.10 & {\bf 0.03} & 0.27 & 1\\
				1 & 0.15 & {\bf 0.10} & 0.51 & 0\\
				2 & 0.06 & {\bf 0.04} & 0.81 & 1\\
				2 & 0.13 & {\bf 0.02} & 0.92 & 0\\
				2 & 0.15 & {\bf 0.10} & 0.92 & 0
			\end{pmatrix}$	
		\end{small}
	\end{center}
	\vskip -0.2in
\end{figure}

\begin{figure}[H]
	\caption{Steps \ref{step:xmap} and \ref{step:tmap} of processing the last table in Figure~\ref{fig:initialsteps}. The candidate split points $\{x^{(1,\ell_1)}\}_{\ell_1}$ for the covariate are set as \{0.51, 0.81\} in this example. Notice that the covariate value for the third row, 0.51, coincides with the smallest candidate split point. In the table on the right, we set it to the smallest value (0.27) observed in the data. This is always possible because under \xgbb's splitting convention, the smallest value will never be chosen as a split point.}
	\label{fig:lattersteps}
	\vskip 0.1in
	\begin{center}
		\begin{small}
			$
			\begin{pmatrix}
				i & \underline\tau & w & \chi & \Delta \\
				\hline\\
				1 & 0.01 & 0.09 & 0.27 & 0\\
				1 & 0.10 & 0.03 & 0.27 & 1\\
				1 & 0.15 & 0.10 & 0.51 & 0\\
				2 & 0.06 & 0.04 & 0.81 & 1\\
				2 & 0.13 & 0.02 & 0.92 & 0\\
				2 & 0.15 & 0.10 & 0.92 & 0
			\end{pmatrix}
			\text{\huge $\rightarrow$}
			\begin{pmatrix}
				i & \underline\tau & w & \chi & \Delta \\
				\hline\\
				1 & {\bf 0} & 0.09 & 0.27 & 0\\
				1 & 0.10 & 0.03 & 0.27 & 1\\
				1 & 0.15 & 0.10 & {\bf 0.27} & 0\\
				2 & {\bf 0} & 0.04 & {\bf 0.51} & 1\\
				2 & {\bf 0.10} & 0.02 & {\bf 0.81} & 0\\
				2 & 0.15 & 0.10 & {\bf 0.81} & 0
			\end{pmatrix}
			$	
		\end{small}
	\end{center}
	\vskip -0.1in
\end{figure}

\section{Preprocessing for invasive ventilation data in MIMIC-IV}
We begin by applying the MIMIC-III preprocessing pipeline to MIMIC-IV in order to extract participants and a commonly available set of variables \citep{harutyunyan2019multitask}. Given the severity of a surgical tracheotomy, a patient is censored at the time of receiving one should this happen.
Following the literature on forecasting iV, we commence risk estimation after 24 hours \citep{clark2013clinical}, resulting in a cohort of 29,108 patients (36,068 ICU stays). In addition to the features extracted by the pipeline (demographics and vital measurements), we add respiratory related measurements that are pertinent to iV, namely, inspiratory/expiratory, oxygen flow and consumption, and respiratory rate.
Due to the functional nature of the data, we carry the last observed value forward for missing value imputation.

The complete list of the extracted variables are as follows:

\begin{center}
\begin{longtable}{lll}
\caption{The selected clinical variables for iV risk estimation in MIMIC-IV.} \label{tab:long} \\

\hline \multicolumn{1}{c}{\textbf{Variable}} & \multicolumn{1}{c}{\textbf{MIMIC-IV table}} & \multicolumn{1}{c}{\textbf{Type}} \\ \hline 
\endfirsthead

\multicolumn{3}{c}%
{{\bfseries \tablename\ \thetable{} -- continued from previous page}} \\
\hline \multicolumn{1}{c}{\textbf{Variable}} & \multicolumn{1}{c}{\textbf{MIMIC-IV table}} & \multicolumn{1}{c}{\textbf{Type}} \\ \hline 
\endhead

\hline \multicolumn{3}{|r|}{{Continued on next page}} \\ \hline
\endfoot

\hline \hline
\endlastfoot

Capillary refill rate & chartevents & categorical \\
Ethnicity & chartevents & categorical \\
Gender & chartevents & categorical \\
Diastolic blood pressure & chartevents & continuous \\
Fraction inspired oxygen & chartevents & continuous \\
Glascow coma scale total & chartevents & continuous \\
Glucose & chartevents, labevents & continuous \\
Heart Rate & chartevents & continuous \\
Mean blood pressure & chartevents & continuous \\
Oxygen saturation & chartevents, labevents & continuous \\
Systolic blood pressure & chartevents & continuous \\
Temperature & chartevents & continuous \\
pH & chartevents, labevents & continuous \\
Age & chartevents & continuous \\
Height & chartevents & continuous \\
Weight & chartevents & continuous \\
Inspired O2 Fraction & chartevents & continuous \\
Respiratory Rate & chartevents & continuous \\
O2 saturation pulseoxymetry & chartevents & continuous \\
PEEP set & chartevents & continuous \\
Inspired Gas Temp. & chartevents & continuous \\
Paw High & chartevents & continuous \\
Vti High & chartevents & continuous \\
Fspn High & chartevents & continuous \\
Apnea Interval & chartevents & continuous \\
Tidal Volume (set) & chartevents & continuous \\
Tidal Volume (observed) & chartevents & continuous \\
Minute Volume & chartevents & continuous \\
Respiratory Rate (Set) & chartevents & continuous \\
Respiratory Rate (spontaneous) & chartevents & continuous \\
Respiratory Rate (Total) & chartevents & continuous \\
Peak Insp. Pressure & chartevents & continuous \\
Plateau Pressure & chartevents & continuous \\
Mean Airway Pressure & chartevents & continuous \\
Total PEEP Level & chartevents & continuous \\
Inspiratory Time & chartevents & continuous \\
Expiratory Ratio & chartevents & continuous \\
Inspiratory Ratio & chartevents & continuous \\
Ventilator Tank \#1 & chartevents & continuous \\
Ventilator Tank \#2 & chartevents & continuous \\
Tidal Volume (spontaneous) & chartevents & continuous \\
PSV Level & chartevents & continuous \\
O2 Flow & chartevents & continuous \\
O2 Flow (additional cannula) & chartevents & continuous \\
Flow Rate (L/min) & chartevents & continuous \\
CO2 production & chartevents & continuous \\
Cuff Pressure & chartevents & continuous \\
ETT Position Change & chartevents & continuous \\
ETT Re-taped & chartevents & continuous \\
Spont Vt & chartevents & continuous \\
Spont RR & chartevents & continuous \\
MDI \#1 Puff & chartevents & continuous \\
Cuff Volume (mL) & chartevents & continuous \\
Trach Care & chartevents & continuous \\
MDI \#2 Puff & chartevents & continuous \\
MDI \#3 Puff & chartevents & continuous \\
Negative Insp. Force & chartevents & continuous \\
Vital Cap & chartevents & continuous \\
BiPap O2 Flow & chartevents & continuous \\
PCV Level & chartevents & continuous \\
BiPap EPAP & chartevents & continuous \\
BiPap IPAP & chartevents & continuous \\
Pinsp (Draeger only) & chartevents & continuous \\
Recruitment Duration & chartevents & continuous \\
PeCO2 & chartevents & continuous \\
Recruitment Press & chartevents & continuous \\
Nitric Oxide & chartevents & continuous \\
Nitric Oxide Tank Pressure & chartevents & continuous \\
Transpulmonary Pressure (Exp. Hold) & chartevents & continuous \\
Transpulmonary Pressure (Insp. Hold) & chartevents & continuous \\
P High (APRV) & chartevents & continuous \\
P Low (APRV) & chartevents & continuous \\
T High (APRV) & chartevents & continuous \\
T Low (APRV) & chartevents & continuous \\
Small Volume Neb Dose \#2 & chartevents & continuous \\
ATC 
BiPap bpm (S/T -Back up) & chartevents & continuous \\
Peak Exp Flow Rate & chartevents & continuous \\
Vd/Vt Ratio & chartevents & continuous \\
\end{longtable}
\end{center}

To track iV recurrence, we also add the cumulative number of past iVs as well as time since last iV (if any).

We randomly split this patient cohort into training and test sets using the approach in \citet{harutyunyan2019multitask} (with modification for MIMIC-IV), resulting in 24,764 patients (30,716 ICU stays) in the training set and 4,344 patients (5,352 ICU stays) in the test set. A 5-fold cross validation is used on the training set to select \softc hyperparameters. The tuned hyperparameters for \softc are \{$100$ trees, depth $3$\}.

\section{Synthetic dataset for scalability analysis}
Letting $X(t)$ be a piecewise-constant covariate with values drawn from $U(0,1]$, we simulate event times from the following hazard function:
\[
\lambda(t,X(t))=B(t,2,2)\times B(X(t),2,2),
t\in (0, 1],
\]
where $B(\cdot, a, a)$ is the PDF of the Beta distribution (with shape and scale $a$). This means that $\lambda$ takes the form of Beta PDFs. If the event has not occurred by $t=1$ the subject is administratively censored at that point. In addition to $X(t)$, we add up to 40 irrelevant covariates to the dataset. The event histories for 5,000 subjects are drawn for training following \citep{wang2020boxhed}. For benchmarking against boosted parametric models, we draw up to 10 million rows of data.
\vskip -5pt
\end{appendix}


\end{document}